\documentclass[noheadings, nojss]{jss}

\usepackage{subfigure}
\usepackage{bm}
\usepackage{amsmath, amsfonts}
\usepackage{setspace}

\author{
David R{\"u}gamer\\ LMU Munich \AND
Chris Kolb \,\,
Cornelius Fritz \,\,
Florian Pfisterer \,\, 
Philipp Kopper \,\, 
Bernd Bischl\\ LMU Munich\\ \AND
Ruolin Shen \,\, Christina Bukas  \,\, Lisa Barros de Andrade e Sousa  \,\, Dominik Thalmeier \\ Helmholtz AI Munich \AND
Philipp F. M. Baumann \\ ETH Zurich \And
Lucas Kook \\ University of Zurich \And
Nadja Klein \\ Humboldt-Universit\"at zu Berlin \AND 
Christian L. M{\"uller}\\ LMU Munich, ICB, Helmholtz Zentrum Munich,\\ CCM, Flatiron Institute, New York}

\title{\pkg{deepregression}: A Flexible Neural Network Framework for Semi-Structured Deep Distributional Regression}

\Plainauthor{David R{\"u}gamer et al.}    
\Plaintitle{deepregression: A Flexible Neural Network Framework for Semi-Structured Deep Distributional Regression} 
\Shorttitle{\pkg{deepregression}: Semi-Structured Deep Distributional Regression} 

\Abstract{
In this paper we describe the implementation of semi-structured deep distributional regression, a flexible framework to learn conditional distributions based on the combination of additive regression models and deep networks. Our implementation encompasses (1) a modular neural network building system based on the deep learning library \pkg{TensorFlow} for the fusion of various statistical and deep learning approaches, (2) an orthogonalization cell to allow for an interpretable combination of different subnetworks, as well as (3) pre-processing steps necessary to set up such models. The software package allows to define models in a user-friendly manner via a formula interface that is inspired by classical statistical model frameworks such as \pkg{mgcv}. The packages' modular design and functionality provides a unique resource for both scalable estimation of complex statistical models and the combination of approaches from deep learning and statistics. This allows for state-of-the-art predictive performance while simultaneously retaining the indispensable interpretability of classical statistical models.}
\Keywords{additive predictors; deep learning; effect decomposition; orthogonal complement; penalization and smoothing}

\Address{
  David R{\"u}gamer\\
  Chair of Statistical Learning \& Data Science\\
  Department of Statistics\\
  LMU Munich\\
  80539, Munich, Germany\\
  E-mail: \email{david.ruegamer@stat.uni-muenchen.de}
}

\begin{document}

\section{Introduction} 

In regression analysis, the main objective is to obtain information about the conditional distribution of a response variable given a set of explanatory variables. Nonetheless, many classical regression approaches only model the mean of this conditional distribution. These mean regression models often cannot adequately account for heteroscedasticity of the conditional distribution or its possible dependence on explanatory variables (or features). As a result, methodological research has brought forward a number of distinct approaches, often referred to as \emph{distribution}, \emph{distributional} or \emph{density regression}. The common ground of all these contributions is to allow the complete set of parameters of the conditional distribution to depend on the available features, not only the mean. Common examples include extensions of generalized additive models \citep[GAMs; see, e.g.,][]{Wood.2017} by making all parameters of a parametric distribution feature-dependent \citep[e.g., generalized additive models for location, scale, and shape (GAMLSS);][]{Rigby.2005}, but also approaches avoiding the parametric distributional assumptions such as Bayesian nonparametrics \citep{Dunson.2010}, 
mixtures of experts \citep{Jacobs.1991}, or conditional transformation models \citep{Hothorn.2014}. Another research branch focuses on directly relating local properties of the conditional distribution, such as quantiles and expectiles, to features \citep[see, e.g.,][]{Koe2005,SchEil2009}. 

Despite being more flexible than mean regression approaches, many of these distributional regression methods are not scalable to either a large number of observations or a large number of features. While these large data settings might already be challenging for mean regression approaches, distributional regression usually involves additional model terms and thus increased computational complexity. The semi-structured deep distributional regression (SDDR) framework proposed by \citet{Ruegamer.2020} and implemented in the package \pkg{deepregression} overcomes these limitations by casting the distributional regression task in a neural network. The package allows the user to learn the distributional parameters of a wide range of parametric densities based on a combination of structured additive regression models and deep neural networks. \pkg{deepregression} thereby offers a scalable and flexible alternative to existing distributional regression approaches. 
A Python-native version of \pkg{deepregression} is implemented in a separate software package \pkg{PySDDR} (\url{https://github.com/HelmholtzAI-Consultants-Munich/PySDDR}).
\\

\subsection{Semi-structured deep distributional regression} \label{sec:sddr}

SDDR is based on ideas from distributional regression for parametric distributions~\citep[see, e.g.,][]{Klein.2015}, to estimate the entire conditional distribution of a continuous, discrete, mixed, or multivariate response variable $\bm{Y}$. This is done through modelling the corresponding distributional parameters. Apart from its application to transformation models \citep{Baumann.2020, atms}, SDDR has also been applied recently to survival analysis \citep{kopper2020, kopper2022deeppamm}, mixture models \citep{Ruegamer.2020b}, and epidemiological models combined with graph neural networks \citep{fritz2021combining}. 

In particular, SDDR constitutes a distributional learning framework in which each distributional parameter $\theta_k \equiv \theta_k(\eta_k), k = 1,\ldots, K$ of an arbitrary pre-specified parametric distribution $\mathcal{D}(\theta_1,\ldots,\theta_K)$ can be modeled through (potentially different) additive predictors $\eta_k$.\footnote{It is assumed that $\mathcal{D}$ has an absolute continuous density with existing finite first derivatives with respect to all distributional parameters.} Each of these predictors is defined as a sum of structured predictors, that can include (penalized) linear effects of features $\bm{x}$, (penalized) smooth effects of features $\bm{z}$, or arbitrary additional (unstructured) neural network predictors for features $\bm{u}$:
\begin{equation}
   \eta_k\equiv \eta_k(\bm{x},\bm{z},\bm{u}) = \bm{x}^\top \bm{w} + \sum_{j=1}^J f_{k,j}(\bm{z}) + \sum_{l=1}^L d_{k,l}(\bm{u}). \label{predtypes}
\end{equation}
The different features $\bm{x}, \bm{u}, \bm{z}$ are also allowed to overlap, giving rise to a generic additive identifiability problem that is handled by our later introduced orthogonalization cell. In \eqref{predtypes}, the first summand on the right-hand side represents the linear effects of features $\bm{x}$ with unknown weights $\bm{w}$, the second part comprises smooth effects $f_{k,j}(\bm{z})$ of one or several features in $\bm{z}$, and the last part is a sum of (deep) neural networks depending on $\bm{u}$. The smooth functions $f_{k,j}$ are defined as linear combinations of appropriate basis functions $B_1,\ldots,B_M$ evaluated at the feature value(s), such as B-spline basis functions~\citep[see][for an overview]{Wood.2017}. Consecutively, $\eta_k, k = 1, ..., K$ is transformed by a one-to-one mapping $h_k:\mathbb{R}\to\Theta_k$ from the real line to the respective parameter space $\Theta_k$ of $\theta_k$ (also known as response functions). For example, choosing $h_k = \exp(\cdot)$ is often used to model a positive scale parameter such as the variance of a Gaussian random variable. In summary, the structural and stochastic components of the model are defined by: 
\begin{align}
    \bm{Y} | \bm{x,z,u} \sim \mathcal{D}(\theta_1,\ldots,\theta_K),
    \label{eq:dist_assumption}
\end{align}
with the following transformations:
$$
(\bm{x},\bm{z},\bm{u}) \overset{\eqref{predtypes}}{\longrightarrow} \eta_k(\bm{x},\bm{z},\bm{u}) \overset{h_k}{\longrightarrow} \theta_k(\eta_k) \longrightarrow \mathcal{D}(\theta_1,\ldots,\theta_k,\ldots,\theta_K).
$$
To ensure identifiability and interpretability in the additive predictors, the SDDR framework uses an orthogonalization cell that allows to separate effects in the structured part from the unstructured part of the predictors. This only requires the neural network predictors $d_l$ to have a fully-connected layer as penultimate layer (cf. Figure~\ref{fig:oz}).

The SDDR framework encompasses many special cases such as linear and generalized linear models, GAMs, and GAMLSS, but also extends those classes to include additional neural network predictors. To allow practitioners to make use of such a flexible framework, \textbf{deepregression} implements the SDDR framework in an accessible and modular manner in the software environment \proglang{R} \citep{R}. While various other packages for distributional regression models exist such as \pkg{gamlss} \citep{Rigby.2005}, \pkg{gamlss.lasso} \citep{Ziel.2021} or \pkg{bamlss} \citep{Umlauf.2019}, to the best of our know\-ledge no software package exists that can arbitrarily combine structured regression models with deep neural networks and thus allow for applications with multimodal data (which is the conventional term in the machine and deep learning community to refer to data sets with different modalities, such as tabular and image data). Before going into the details of our package, we introduce our running example used throughout this tutorial.

\subsection{Case study: Airbnb price listing data}\label{sec-case1}

Airbnb is a company providing an online marketplace for rental apartments. In this example, our goal is to construct a model to predict the listing price of apartments with all available information. From the website \url{http://insideairbnb.com}, we obtain all current listings of apartments on the Airbnb website. The provided information includes various data modalities, ranging from numeric variables, such as latitude and longitude, integer variables, such as the number of bedrooms, textual information, like a room description, dates, to a picture of each property. This breadth of different data types makes the dataset an ideal candidate for the application of \pkg{deepregression}. The final model could then, e.g., be deployed as a guide for new costumers wanting to rent an apartment in order to help setting a reasonable price for their own advertisement.  In order to reduce computational complexity, we will focus on apartments in Munich (Germany) and show example code for demonstration purposes.

The cleaned \texttt{airbnb} data from Munich consist of 3,504 apartment listings. In order to demonstrate the functionality of our package, we will focus on a handful of features, namely: \code{description} (a textual description of the property), \code{host\_since} (a date variable describing the host's membership duration), \code{longitude} and \code{latitude}, \code{review\_score\_value} (a review score), \code{host\_response\_time} (the time until the host response to inquiries), and the properties' images (which are named according to the \code{id} variable in the dataset). The response variable is the price of the respective apartment \code{price}. 

\begin{verbatim}
R> airbnb <- readRDS("munich.RDS")
R> airbnb$days_since_last_review <- as.numeric(
+    difftime(airbnb$date, airbnb$last_review)
+  )
\end{verbatim}

The remainder of this paper is structured as follows:  Section~\ref{sec:rpack} details the core functions and building blocks in the corresponding \proglang{R} package. A brief summary and conclusion is given in the final Section~\ref{sec:conclusion}.

\section{The package} \label{sec:rpack}

\begin{figure}
\hspace*{-0.3cm}    \includegraphics[width=1\textwidth]{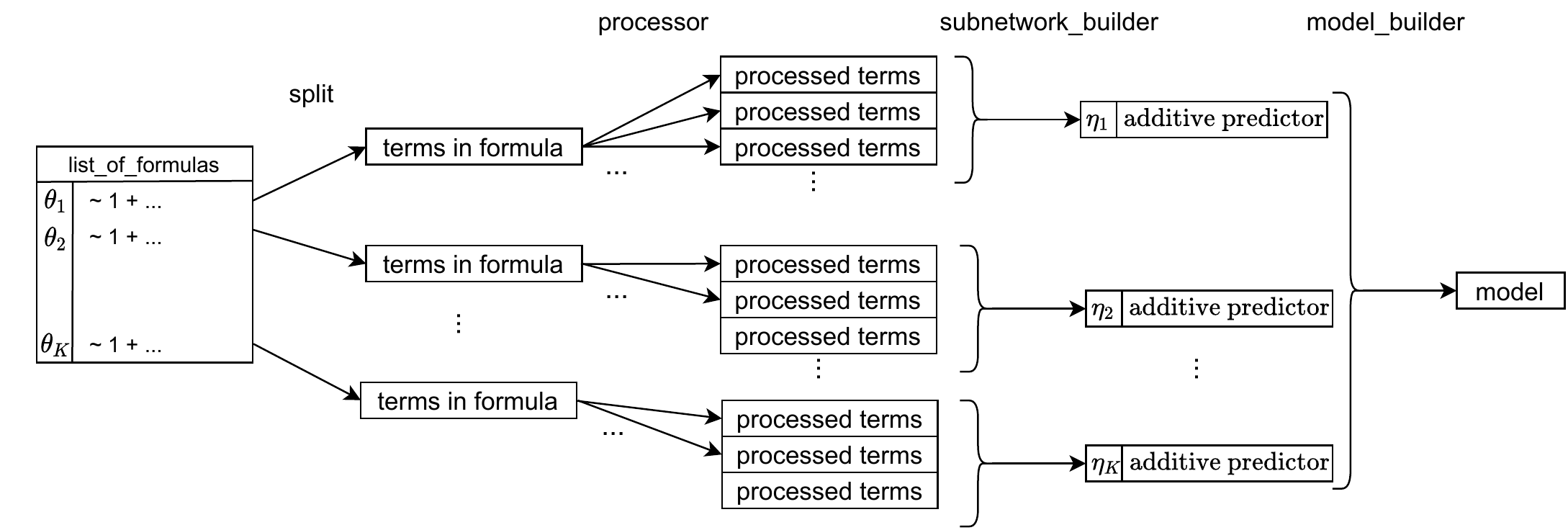}
    \caption{Model building steps in the package. The list of formulas provided by the user is first split up and each additive term is processed by its respective processor. Thereafter, all processed terms of one formula entry are combined to an additive predictor using the a subnetwork builder function. Results are then combined to form the final model using the a model builder function.}
    \label{fig:sequence}
\end{figure}

\pkg{deepregression} is constructed in a modular fashion, allowing to flexibly specify a large number of models as a sequence of processing steps, which are illustrated in Figure~\ref{fig:sequence}. In the first step, the provided formulas defining all additive predictors are parsed using \code{processor} functions. Every formula term (linear term, smooth term, ...) is handled by its respective \code{processor}. The processed terms each have their own methods for providing data for model fitting and prediction, as well as functions to return the coefficients and a plot of the partial effect if appropriate. The parsed formulas containing all these processed terms are returned to \pkg{deepregression}. A second, larger processing step constructs the subnetworks for each additive predictor $\eta_k, k = 1, ..., K$. These subnetworks are finally passed to a model builder function that combines all additive predictors into a model. Per default, model building and fitting is done using \pkg{keras}, but \pkg{deepregression} allows to replace this estimation engine by another optimization routine, e.g., using lower-level \proglang{TensorFlow} functions.

\subsection{Core functionality}

\begin{figure}[ht!]
     \begin{center}
        \subfigure[A structured linear network: linear combination of input features $x_j$ with optional $L_1$- or $L_2$-penalty on the weights $w_{k,j}$]{%
            \label{fig:first}
            \includegraphics[page=1,width=0.36\textwidth]{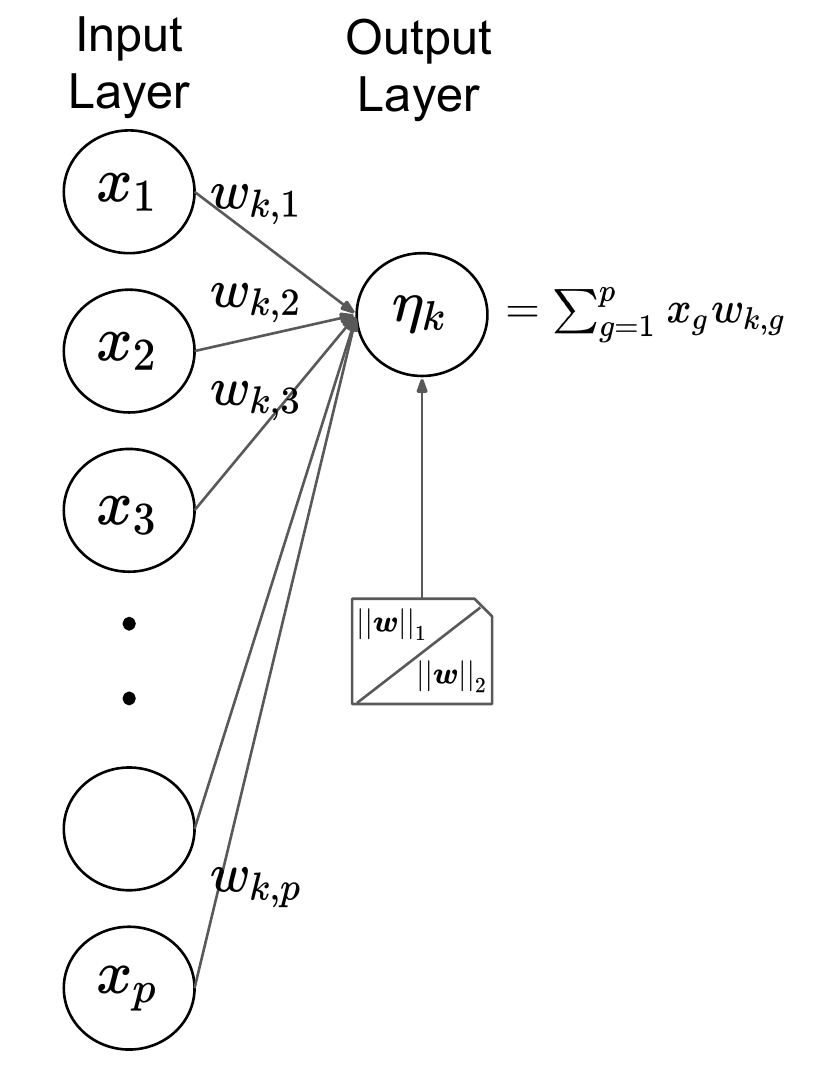}
        }%
        \quad
        \subfigure[A structured non-linear network: linear combination of (basis-transformed) features $B_m(z_1)$ via $M$ basis functions $B_m, m=1,\ldots,M$ with optional matrix-based quadratic penalty $\bm{P}$ on the weights $\bm{\gamma}=(\gamma_1,\ldots,\gamma_M)^\top$.]{%
           \label{fig:second}
           \includegraphics[page=2,width=0.56\textwidth]{deepregression_paper_cropped}
        }\\
                \subfigure[A distributional layer learned by $K$ subnetworks: each parameter $\theta_k$ of  $\mathcal{D}$ with density $\mathfrak{p}_{\mathcal{D}}$ can be learned using (a), (b) and/or a DNN.]{%
            \label{fig:third}
            \includegraphics[page=3,width=0.7\textwidth]{deepregression_paper_cropped}
        }

    \end{center}
    \caption{Illustration of network components that can be combined using the package. DNNs are left out as these can be arbitrarily specified and combined with additive predictors as long as the penultimate layer in the DNN corresponds to a fully-connected layer.}%
   \label{fig:subfigures}
\end{figure}

\pkg{deepregression}'s core functionality is to define and fit (deep) distributional regression models. In the package, various parametric response distributions can be defined, and their parameters structurally related to the three different additive predictor types defined in \eqref{predtypes}. For example, a three-parameter distribution, such as the location, scale, and shape-parameterized t-distribution, can have different predictors defined for each of the $K=3$ parameters. Here, the scale and shape parameter could be nuisance parameters that are modeled using a deep neural network (DNN) of all features, while the location parameter is assumed to have specific (non-)linear effects in certain features and is thus modeled in a structured manner. From a network perspective, the main building blocks are
(i) a structured linear layer,
   (ii) a structured non-linear layer,
    (iii) distributional layer(s), and 
    (iv) an orthogonalization cell.
Building blocks (i) - (iii) are visualized in Figure~\ref{fig:subfigures} and (iv) will be discussed in more detail later.  
The distributional layer, illustrated in Figure~\ref{fig:third}, defines the final distribution using one or more subnetworks for each parameter $\theta_k$. The network is then trained using the negative log-likelihood resulting from \eqref{eq:dist_assumption} as a loss function. In each subnetwork, the additive predictors $\eta_k$ can be defined using (penalized) linear effects, Figure~\ref{fig:first}, (penalized) smooth effects, Figure~\ref{fig:second}, or arbitrary deep neural networks. 

Furthermore, the distributional layers can be combined to define a richer distribution family, e.g., to create zero-inflated distributions or mixtures of distributions.

While the structured linear layer can take inputs in their original form, the structured non-linear layer requires additional pre-processing which is described below as part of the three core functionalities. The DNNs for the unstructured effects $d_{k,l}(\bm{u})$ have to be user-specified to correctly process the inputs.

\subsection{Main components}

\subsubsection{Formula interface}

In the formula interface, each distributional parameter can be specified by a symbolic additive predictor. A model with two parameters $\theta_1, \theta_2$ for location and scale of a distribution $\mathcal{D}$ with $\theta_1 = w_0 + w_1 x + f_{1,1}(z_1)$ and $\theta_2 = f_{2,1}(z_2) + d_{2,1}(\bm{u})$ in \proglang{R} could translate to:

\newpage

\begin{verbatim}
R> list_of_formulas = list(
+    loc = ~ 1 + x + s(z1, bs="tp"),
+    scale = ~ 0 + s(z2, bs="ps") + dnn(u)
+  )
\end{verbatim}
This code snippet specifies an intercept (also called bias term) for the location parameter, a linear effect for $x$, and a thin-plate regression spline \citep{Wood.2003} for $z_1$. The scale parameter is specified without bias term, but using another non-linear effect of feature $z_2$, now characterized by a penalized spline. It further contains a DNN predictor for inputs $\bm{u}$, where \texttt{dnn} is a model defined separately with suitable input shape. In the network, this formula will create two subnetworks, both with one output unit (the predictors $\eta_k, k=1,2$ for the location and scale parameters), which are then used to define the distribution.\\

\textbf{Case study: formula interface} 

We start with a simple additive model to predict the log-transformed price 
\begin{verbatim}
R> y = log(airbnb$price)
\end{verbatim}
using a normal distribution parameterized by a location and scale parameter. We learn the location using an intercept and a tensor-product spline for \code{longitude} and \code{latitude}, while estimating the scale to be constant. In our package, this corresponds to
\begin{verbatim}
R> list_of_formulas = list(
+    loc = ~ 1 + te(latitude, longitude, df = 5),
+    scale = ~ 1
+  )
\end{verbatim}

\subsubsection{Processors}

As described in Figure~\ref{fig:sequence}, each additive term contained in predictor $\eta_k,k=1,\ldots,K$ is handled by type-specific processors. \pkg{deepregression} comes with different processors that automatically parse each formula. These are listed in Table~\ref{tab:processors}. It also allows the user to supply custom processors using the argument \code{additional_processors}. 

\begin{table}[]
\begin{tabular}{l|lp{3.8cm}}
 & Notation & Arguments \\ \hline
Intercept & \code{1} & - \\
Linear term & \code{x} or \code{lin(x)} & -  \\
Ridge-penalized linear effects & \code{ridge(x, la = ...)}  &  penalty \code{la}\\
Lasso-penalized linear effects & \code{lasso(x, la = ...)}  &  penalty \code{la}\\
Offset & \code{offset(x)}  & -  \\
(Tensor-product) Splines & \code{s}/\code{te}/\code{ti(x, ..., df = ...)} & degrees-of-freedom \code{df}; \pkg{mcgv} arguments  
\end{tabular}
\caption{Processors for formula terms provided in the package.}
\label{tab:processors}
\end{table}

\subsubsection{Network initialization}

After defining the model formulas for each distributional parameter, the SDDR model is initialized via the \code{deepregression} function. The required arguments are the response \code{y}, the data frame or list \code{data} including all features, the distribution \code{family}, the \code{list_of_formulas}, and the \code{list\_of\_deep_models} to define the (deep) neural networks that are used to model unstructured effects.\\

\textbf{Case study: network initialization} 

\begin{verbatim}
R> mod_simple <- deepregression(
+    y = y, 
+    data = airbnb, 
+    list_of_formulas = list_of_formulas,
+    list_of_deep_models = NULL
+  )
\end{verbatim}

The result \code{mod\_simple} is of class \code{deepregression}, a list of length three, where the first entry \code{model} is a \pkg{keras} model, the second element \code{init\_params} is a list of objects required for model fitting, tuning, plotting, and other post-processing functionalities and the third element \code{fit\_fun} encompassing the fitting function needed for model fitting. The \code{model} element contains the references to the respective Python objects and is independent of the actual data. Since \code{keras} models do not store any data, the second list item \code{init\_params} is returned for convenience to give the user direct access to the pre-processed data and to simplify calls to other methods such as \code{fit}.

\begin{verbatim}
R> class(mod_simple$model)
[1] "keras.engine.functional.Functional"                        
[2] "keras.engine.training.Model"                               
[3] "keras.engine.base_layer.Layer"                             
[4] "tensorflow.python.module.module.Module"                    
[5] "tensorflow.python.training.tracking.tracking.AutoTrackable"
[6] "tensorflow.python.training.tracking.base.Trackable"        
[7] "keras.utils.version_utils.LayerVersionSelector"            
[8] "keras.utils.version_utils.ModelVersionSelector"            
[9] "python.builtin.object"   
\end{verbatim}

\subsubsection{Pre-processing for structured non-linear layers}

Structured non-linear effects can be fitted in \pkg{deepregression} by defining a corresponding pre-processing function that generates a design matrix based on (user-defined) basis functions and a penalty matrix that is used to regularize the smoothness of the estimated non-linear functions. In R, these objects are computed by calling the function \texttt{smooth.construct} from the package \pkg{mgcv} \citep{Wood.2001}.\\

\textbf{Case study: Pre-processing} 

In the above defined model \code{mod}, the smooth terms from \pkg{mgcv} are stored in the \code{init\_params} element and are used for model fitting, prediction, and plotting.

\begin{verbatim}
R> sapply(mod_simple$init_params$parsed_formulas_contents$loc,
  "[[", "term")
[1] "te(latitude, longitude)" "(Intercept)"
\end{verbatim}

\subsubsection{Specification of the family}

The \code{family} argument specifies the respective family to be learned. Possible choices can be searched for in \code{?dr\_families}, comprising over 30 different options, including those from an exponential family (such as Gaussian, Bernoulli, Poisson, beta, gamma, and multinomial distribution).

\pkg{deepregression} can also define custom distributions. The add-on \proglang{R} package \pkg{mixdistreg} further allows to fit mixtures of families, as proposed by \citet{Ruegamer.2020b}. Details can be found in the Section~\ref{sec:advanced-usage}.

\subsubsection{Formula contents} \label{subsec:formula}

The formulas defining each distributional parameter can be specified in the same way as in \code{mgcv::gam}, including \code{s}-terms, \code{ti}- and \code{te}-terms that can be further customized by using different basis functions, penalty terms, basis dimensions or knot placement (see \code{?mgcv::s}). Factor variables in the formula are also treated the same way as in conventional regression packages by automatically transforming them to dummy-coded features. The exclusion of the intercept in a linear predictor can be defined as per usual by adding the \code{0} or \code{-1} to the formula. Table~\ref{tab:processors} lists all processors for structured terms that are supported by default.

\subsubsection{Specification of the DNNs} \label{sec:specification-of-the-dnns}

The DNNs specified in the \code{list\_of\_formulas} must also be passed as named list elements in the \code{list\_of\_deep\_models}. The named list \code{list\_of\_deep_models} contains a list of DNNs, each specified as a function using the \code{keras} layers API. The respective functions take as many inputs as defined for the neural network and have a fully-connected output layer containing one hidden unit (or any other layer that results in the appropriate amount of outputs as discussed later). For instance, an example DNN for the features $\bm{u}$ from the previous example can be specified using the pipe \code{\%>\%} notation as follows:

\begin{verbatim}
R> deep_model <- function(x)
+    {
+      x %>% 
+      layer_dense(units = 5, activation = "relu", use_bias = FALSE) %>%
+      layer_dropout(rate = 0.2) %>%
+      layer_dense(units = 3, activation = "relu") %>%
+      layer_dropout(rate = 0.2) %>%
+      layer_dense(units = 1, activation = "linear")
+    }
\end{verbatim}

\code{deep\_model} defines a two hidden-layer DNN with $20\%$ dropout between each layer and ReLU activation functions. During training the DNN randomly masks $20\%$ of the weights for every observation for these layers due to the specified dropout in order to regularize the network. Since the DNN's output enters the real-valued predictor $\eta_k$ additively, it is typically most sensible to use a \code{'linear'} activation function in the output layer for maximum flexibility in the values $\eta_k$.
To ensure identifiability of the structured model predictor when structured regression terms and a DNN share some input features, an orthogonalization cell is automatically included before the last dense layer. In this case, it is required to use a \code{'linear'} activation function as in this example.\\

\textbf{Case study: formulas and DNNs} 

The following code shows an illustrative specification including possible formula terms. We only define the \code{location} using an intercept, categorical effects for beds \code{beds}, two splines (a P-spline and thin-plate regression spline) for the number of guests \code{accommodates} and days since last review \code{days\_since\_last\_review}, as well as a DNN for the review score \code{review\_scores\_rating} and the number of reviews per month \code{reviews\_per\_month}. To make the intercept identifiable, we set an additional option that is explained in detail in Section~\ref{sec:oz}.

\begin{verbatim}
R> options(identify_intercept = TRUE)
R> mod <- deepregression(
+    y = y, 
+    data = data,
+    list_of_formulas = list(
+      location = ~ 1 + beds + s(accommodates, bs = "ps") +
+        s(days_since_last_review, bs = "tp") + 
+        deep(review_scores_rating, reviews_per_month),
+      scale = ~1),
+    list_of_deep_models = list(deep = deep_model)
+ )
\end{verbatim}

\subsection{Model training and tuning}

\subsubsection{Model fitting}

Once the model \code{mod} has been set up, the neural network can be trained using the \code{fit} function. Per default, the \code{fit} function is a wrapper of the corresponding \code{fit} function for \pkg{keras} models and inherits the \code{keras::fit} arguments. More specifically,
\begin{itemize}
\item \code{epochs} to specify the number of iterations,
\item \code{callbacks} to specify information that is called after each epoch or batch (used, e.g., for early stopping),
\item \code{validation\_split} to specify the share of data (in [0,1)) that is used to validate the model (while the rest is used to train the model),
\item \code{validation\_data} to optionally specify any predefined validation data, 
\item \code{verbose} to define if progress is printed in the console,
\item \code{view_metrics} to activate an interactive plot window when using RStudio.
\end{itemize}
Several other arguments are available, such as a logical argument \code{early\_stopping} to activate early stopping and \code{patience} to define the patience used in early stopping.\\

\textbf{Case study: model fitting} 

\begin{verbatim}
R> mod %>% fit(
+    epochs = 100, 
+    verbose = FALSE, 
+    view_metrics = FALSE,
+    validation_split = 0.2
+    )
\end{verbatim}

To inspect the fitted model, we can either use \code{coef} or \code{plot} for structured effects, as will be demonstrated later, or extract the fitted values of the model using \code{fitted} (which yields the prediction on the training data):

\begin{verbatim}
R> fitted_vals <- mod %>% fitted()
R> cor(fitted_vals, y)
          [,1]
[1,] 0.5762063
\end{verbatim}

\subsubsection{Model tuning}

In a similar manner as \code{fit}, \pkg{deepregression} offers a cross-validation function \code{cv} for model tuning. This can be used to fine-tune the model by, e.g., changing the specifications of the formula(s), changing the DNN structure, or determining the amount of smoothness (using the \code{df} argument in \code{deepregression}; see the next section for details). Folds in the \code{cv} function can be specified using the \code{cv\_folds} argument. This argument takes either an integer for the number of folds or a list where each entry is again a list of two, one element with data indices for training and one with indices for testing. The procedure then uses these folds to refit the model the amount of times specified by the integer or the length of the fold list. The \code{patience} (number of epochs with no improvement after which training will be stopped) for early stopping can be set using the respective argument.\\

\textbf{Case study: cross-validation} 

As an example we use 100 iterations and three folds. The results of the cross-validation can be plotted using \code{plot_cv}.

\begin{verbatim}
R> mod_cv <- deepregression(
+    y = y, 
+    data = data,
+    list_of_formulas = list(
+      location = ~ 1 + beds + s(accommodates, bs = "ps") +
+        s(days_since_last_review, bs = "tp") + 
+        deep(review_scores_rating, reviews_per_month),
+      scale = ~1),
+    list_of_deep_models = list(deep = deep_model)
+ )
R> res_cv <- mod_cv %>% cv(
+    plot = FALSE, 
+    cv_folds = 3,
+    epochs = 100
+ )
R> plot_cv(res_cv)
\end{verbatim}

\begin{figure}
    \centering
    \includegraphics{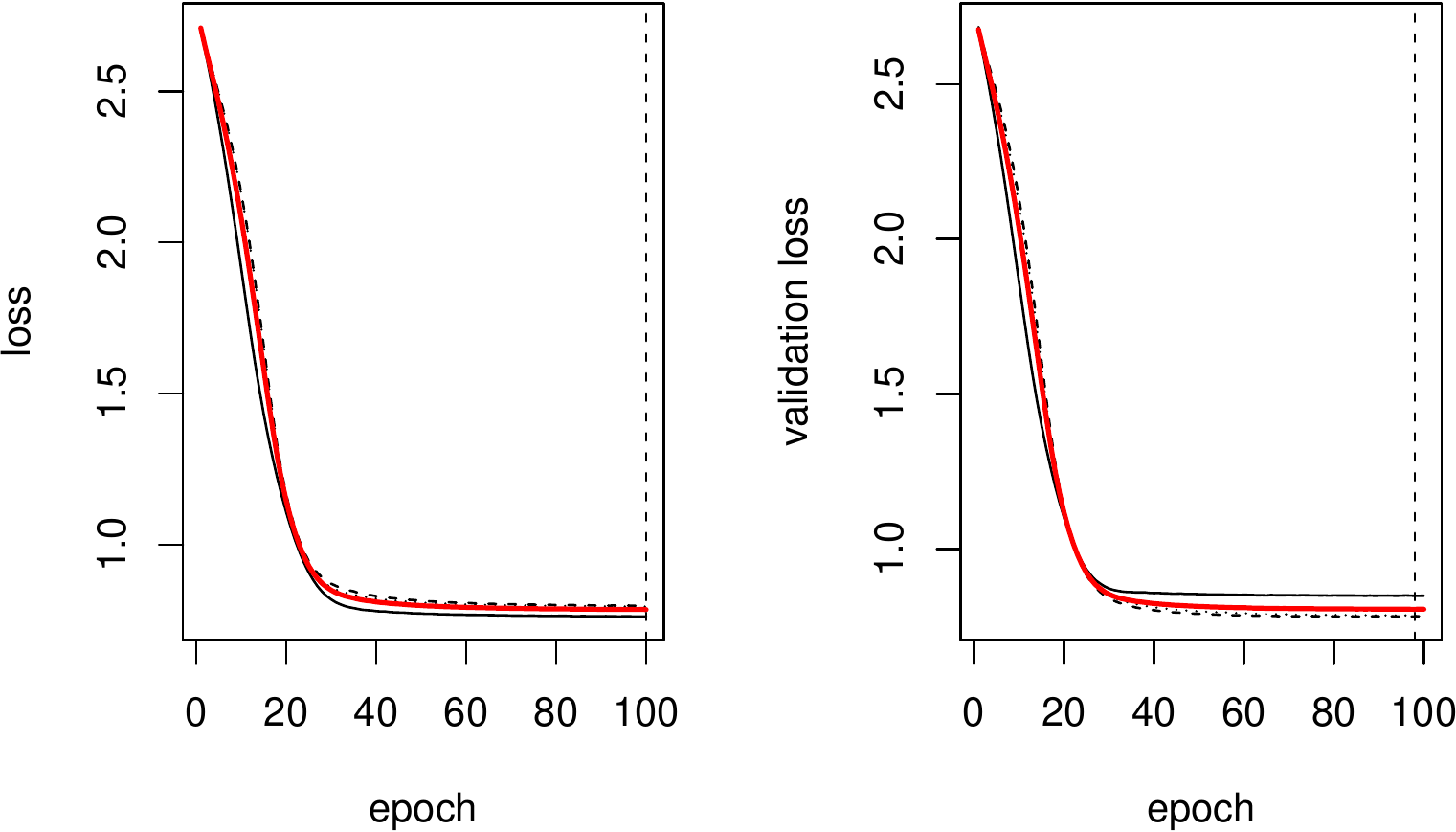}
    \caption{Visualization of cross-validation result. Horizontal black lines indicate the different folds over different epochs (x-axis) and the corresponding loss (left) or validation loss (right). The red line shows the mean loss and validation loss over all iterations. The dashed vertical line indicates the epochs with the lowest average loss and validation loss in the respective plots.}
    \label{fig:my_label}
\end{figure}

\subsection{Methods overview}\label{sec:methods}

Apart from \code{fit} and \code{cv}, there are additional methods that can be applied to \code{deepregression} objects:

\begin{itemize}
    \item \code{coef}: The \code{coef} function extracts the coefficients (network weights) of all structured layers, i.e., coefficients of linear of smooth effects. Using the argument \code{type}, the user can specify what type of coefficients to return (possible choices: \code{"linear"} for linear effects and \code{"smooth"} for basis coefficients of the specified smooth terms). The argument \code{params} allows to select for which of the distributional parameters the coefficients should be returned.
\item \code{plot}: The \code{plot} function can be applied to \code{deepregression} objects to visualize the learned non-linear effect curves for splines and tensor product splines. Using \code{which}, a specific effect can be selected using the corresponding integer in the structured part of the formula (if \code{NULL} all effects are plotted), while the integer given for \code{which_param} indicates which distributional parameter is chosen for plotting. Via the argument \code{only_data = FALSE}, the function can also be used to just return the data used for plotting.
\item \code{predict}: The \code{predict} function works as for other \code{keras} models. It either just takes the model as input and returns the predictions (per default the distribution's expectation) for observations in the training dataset, or, when supplied with new data, produces the corresponding predictions for the new data. As \code{deepregression} learns a distribution and not only its mean, the user can choose what type of distribution characteristic is to be predicted via the argument \code{apply\_fun}, chosen from the distribution methods supplied by \pkg{tfprobability} (per default \code{apply\_fun = tfd\_mean} predicts the mean of the fitted distribution).
\item \code{mean, stddev and quant}: These functions are convenience functions that directly return the mean, standard deviation, or certain quantiles of the learned distribution. All three functions work for the given training data as well as newly provided \code{data}. The \code{quant} function has an additional argument \code{probs} to provide the quantile(s) of interest.
\item \code{get\_distribution}: Instead of returning summary statistics, \code{get_distribution} returns the whole \pkg{TensorFlow} distribution with all its functionalities. The distribution can, e.g., be used to sample random numbers from it or compute the PDF or CDF. 
\item \code{log\_score}: The \code{log_score} function directly returns the evaluated log-likelihood based on the estimated parameters and the provided data (including a response vector \code{this_y}). If \code{data} and \code{this_y} are not provided, the function will calculate the score on the training data.
\end{itemize}

\textbf{Case study: convenience functions} 

We first inspect the estimated linear effects.

\begin{verbatim}
R> coef(mod, type="linear")
$location
 (Intercept)       beds.1       beds.2       beds.3       beds.4       beds.5 
 1.975234151  0.008926539  0.070396960  0.138633251  0.151451677  0.582612634 
      beds.6       beds.7       beds.8       beds.9      beds.10      beds.11 
 0.221924141  0.175573707 -0.311866581 -0.123485431  0.524103165  1.533515930 
     beds.12      beds.16      beds.18      beds.20      beds.30 
-0.313976169 -0.611083746  1.997076750 -0.692837417 -0.524448991 

$scale
(Intercept) 
 -0.6014107 
\end{verbatim}

\begin{figure}
    \centering
    \includegraphics[scale=0.7]{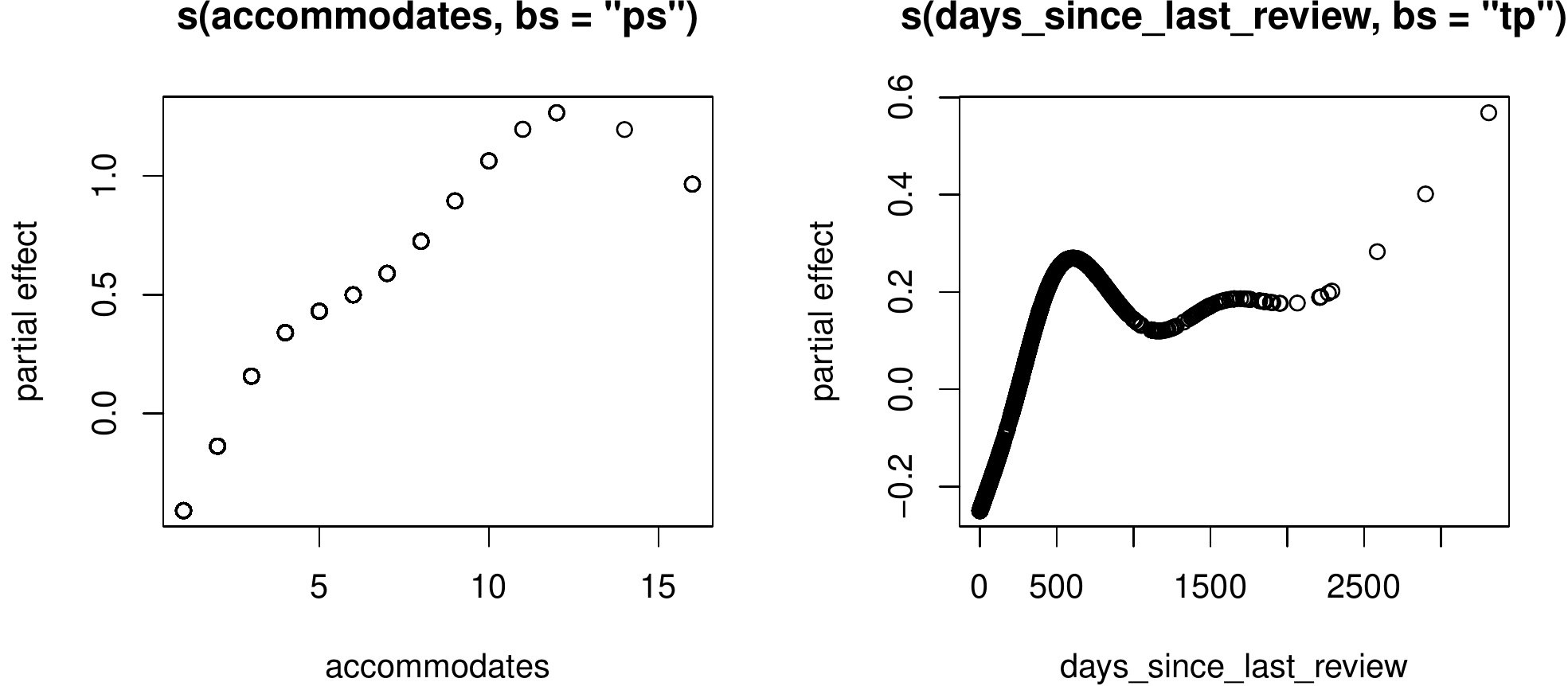}
    \caption{Estimated partial effects of accommodates and the days since the last review on the average log-price of an apartment.}
    \label{fig:plot_mod}
\end{figure}

For non-linear effects, the estimated smooth functions can be plotted using \code{plot} with result visualized in Figure~\ref{fig:plot_mod}.

While most of the other convenience functions are wrappers that access the underlying \pkg{keras} model and convert the quantity of interest to an \proglang{R} object, \code{get\_distribution} returns the fitted distribution, a Python \code{tfp.distributions}.

\begin{verbatim}
R> dist <- mod %>% get_distribution()
R> str(dist, 1)
tfp.distributions.Normal("model_1_distribution_lambda_1_Normal", 
batch_shape=[3504, 1], event_shape=[], dtype=float32)
\end{verbatim}

Instead of working with the original distribution object, the user can work with convenience functions to get the quantity of interests directly, as demonstrated below.

\begin{verbatim}
R> first_obs_airbnb <- as.data.frame(airbnb)[1,,drop=F]
R> dist1 <- mod %>% get_distribution(first_obs_airbnb)
R> meanval <- mod %>% mean(first_obs_airbnb) %>% c
R> q05 <- mod %>% quant(data = first_obs_airbnb, probs = 0.05) %>% c
R> q95 <- mod %>% quant(data = first_obs_airbnb, probs = 0.95) %>% c
R> xseq <- seq(q05-1, q95+1, l=1000)
R> plot(xseq, sapply(xseq, function(x) c(as.matrix(dist1$prob(x)))), 
  type="l", ylab = "density(price)", xlab = "price")
R> abline(v = c(q05, meanval, q95), col="red", lty=2)
\end{verbatim}

\begin{figure}[!h]
    \centering
    \includegraphics[width=0.65\textwidth]{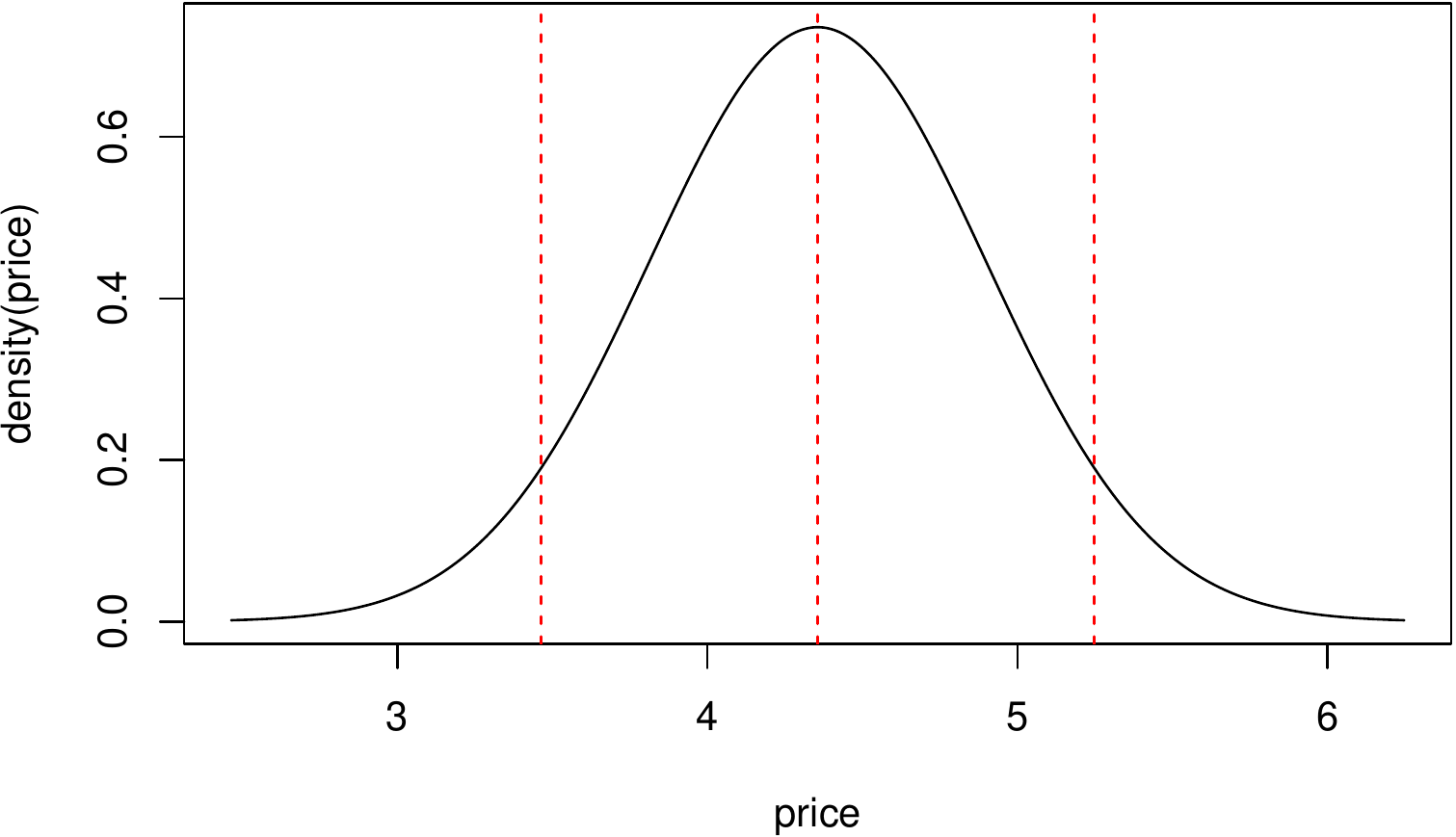}
    \caption{Visualization of the estimated distribution (black line) as well as the mean and quantiles (red vertical lines) for the first data point in the dataset.}
    \label{fig:quantmean}
\end{figure}

Figure~\ref{fig:quantmean} shows the density of the estimated distribution together with estimated mean and quantiles.

\subsection{Penalties}

\pkg{deepregression} allows for different penalties, including $L_1$-, $L_2$-, and other smoothing penalties as provided by the \pkg{mgcv} package. While the latter is implicitly created when using \code{s}-, \code{ti}- or \code{te}-terms in the formula, the $L_1$- and $L_2$-penalties are used to penalize linear effects by defining the amount of penalization via the \code{la} argument in the model terms \code{lasso(...}) \citep[using ideas from][to allow for feature sparsity when using stochastic gradient optimization routines]{hoff2017lasso, Tibs.2021} and  \code{ridge(...)}. Since the model object returned by \code{deepregression} is a list where the first element is a \code{keras} model, additional custom penalties can be specified. This is implemented by the \code{additional\_penalty} argument, which requires a function with one (unused) dummy argument and defines the penalty based on the \pkg{keras} model's \code{trainable\_weights}. Due to mini-batch training, the scaling and influence of penalties is not invariant w.r.t. the data size. Per default, \pkg{deepregression} therefore defines a scaling value \code{sp_scale} that multiplies the penalty by the inverse of the training data size. This behaviour an be adapted through the \code{penalty_options}.\\

\textbf{Case study: additional penalties} 

To access the model's weights, we can access the \code{keras} object in the previously created \code{deepregression} model \code{mod\_simple}, use \code{mod\_simple$model$trainable\_weight} and, e.g., penalize all coefficients (weights) of the defined tensor product spline using an $L_1$-penalty with $\lambda=0.5$:

\begin{verbatim}
R> lambda <- 0.5
R> l1part <- tf$reduce_sum(tf$abs(mod_simple$model$trainable_weights[[1]]))
R> addpen <- function(x) lambda * l1part
R> mod_simple_pen <- deepregression(
+    y = y, 
+    data = airbnb, 
+    list_of_formulas = list_of_formulas,
+    list_of_deep_models = NULL, 
+    additional_penalty = addpen
+  )
\end{verbatim}

Note that the internal function must return a scalar value, and mathematical operations should be specified using \pkg{TensorFlow} functions, available through \code{tf\$...} (and loaded together with \pkg{deepregression}).

\subsubsection{Smoothing penalties}

Apart from specifying the smoothness of \pkg{mgcv}'s smooth terms in the formulas manually (see \code{mgcv::s} for more details), two further options are available in \pkg{deepregression}. The first option is to use a single degrees-of-freedom specification using the argument \code{df} in \code{penalty_control}, a function that defines the \code{penalty_options} argument in \code{deepregression}. Using Demmler-Reinsch orthogonalization \citep[see, e.g.,][]{Ruppert.2003}, all smoothing parameters are then calculated based on this specification. This ensures that no smooth term has more flexibility than the other term. When \code{df} is left unspecified, the default \code{df} is 10. The second option in \code{deepregression} is to specify \code{df} for each smooth term in each formula by providing writing \code{df = ...} into the \code{s/te/ti} term. 

\paragraph{Details} The definition of the degrees-of-freedom can be changed using the \code{hat1} argument in \code{penalty_control}. When set to \code{TRUE}, the \code{df} are defined to be the sum of the diagonal terms (trace) of the hat matrix $\bm{H}$ of the corresponding smooth term. The default is \code{FALSE}, yielding the more common definition of the effective degrees-of-freedom $\text{df} = \text{trace}(2\bm{H} - \bm{H}\bm{H})$. 
As the penalty in \pkg{TensorFlow} is added to each single observation, the penalty term is per default divided by the number observations, preventing too strong penalization of the log-likelihood. In certain situations, it might be necessary to scale the penalty differently. This can be done using the argument \code{sp\_scale} in \code{penalty_control}, which is \code{1/NROW(y)} by default to account for batch training. If training is done in full batch mode (batch size is equal to the total number of observations), we recommend setting \code{sp\_scale} to \code{1}.\\

\textbf{Case study: smoothing penalties} 

In the following we use a single smooth term to demonstrate the effect of \code{df}.

\begin{verbatim}
R> form_df_3 <- list(loc = ~ 1 + s(days_since_last_review, df = 3),
+  scale = ~ 1)
R> form_df_6 <- list(loc = ~ 1 + s(days_since_last_review, df = 6),
+  scale = ~ 1)
R> form_df_10 <- list(loc = ~ 1 + s(days_since_last_review, df = 10),
+  scale = ~ 1)
args <- list(y = y, data = airbnb, 
             list_of_deep_models = NULL)

R> mod_df_low <- do.call("deepregression", 
+  c(args, list(list_of_formulas = form_df_3)))
R> mod_df_med <- do.call("deepregression", 
+  c(args, list(list_of_formulas = form_df_6)))
R> mod_df_max <- do.call("deepregression", 
+  c(args, list(list_of_formulas = form_df_10)))

R> mod_df_low %>% fit(epochs = 2000, early_stopping = TRUE, verbose = FALSE)
R> mod_df_med %>% fit(epochs = 2000, early_stopping = FALSE, verbose = FALSE)
R> mod_df_max %>% fit(epochs = 1000, early_stopping = FALSE, verbose = FALSE)

R> par(mfrow=c(1,3))
R> plot(mod_df_low)
R> plot(mod_df_med)
R> plot(mod_df_max)
\end{verbatim}

The different amount of smoothness resulting from the different choices in degrees-of-freedom is depicted in Figure~\ref{fig:my_label}. 

\begin{figure}
    \centering
    \includegraphics[width=\textwidth]{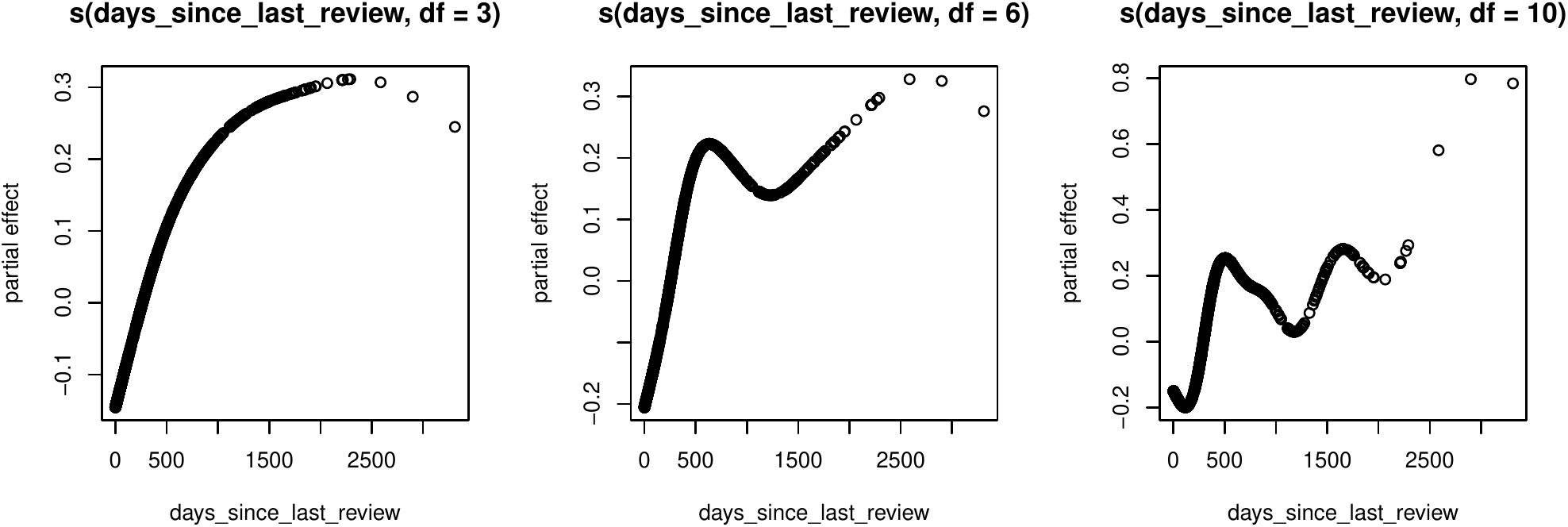}
    \caption{Case study. Comparison of different degrees-of-freedom settings for a single smooth term model.}
    \label{fig:my_label}
\end{figure}

\subsection{Neural network settings}

Since \pkg{deepregression} constitutes a holistic neural network, certain settings and advantages from deep learning can be made use of.

\subsubsection{Shared DNN}

In addition to the DNN specifications introduced in the Section~\ref{sec:specification-of-the-dnns}, \pkg{deepregression} allows the user to share one DNN between some or all distributional parameters. This can drastically reduce the number of parameters to be estimated. The following example will demonstrate how to share a DNN between two parameters. The number of output units has to  be equal to the number of parameters learned in this case. Sharing a network requires its inclusion in the \code{list\_of\_deep\_models} and an additional list with formulas for the argument \code{train\_together}. Analogously to \code{list\_of\_formulas}, this list provides the information which networks are used for which distributional parameters. \\

\textbf{Case study: Shared DNN} 

In the following example, we will learn both location and scale parameter of the rooms' listing prices using an embedding of words contained in the room description. We use the already tokenized text contained in the matrix \code{texts} in the \code{airbnb} data. The network we use consists of an embedding layer that processes the \code{1000} different tokenized words and learns a vector representation of the words in a \code{100}-dimensional space. This representation is then reduced in the second dimension using the mean. Following this operation in the so-called Lambda layer, we use two fully-connected layers with 20 and two units and dropout in between. The two units in the last layer correspond to the two distributional parameters that are learned from the text information. 

\begin{verbatim}
R> embd_mod <- function(x) x %>%
+    layer_embedding(input_dim = 1000,
+      output_dim = 100) %>%
+    layer_lambda(f = function(x) k_mean(x, axis = 2)) %>%
+    layer_dense(20, activation = "tanh") %>% 
+    layer_dropout(0.3) %>% 
+    layer_dense(2) 
\end{verbatim}

In addition to the DNN, we will only use intercepts in both distributional parameters. To not copy and paste the \code{embd\_mod} into every parameter formula, we create another list item which is provided to the \code{list\_of\_formulas}. The \code{mapping} argument then tells \pkg{deepregression} which formula is used for which parameter (here \code{list(1, 2, 1:2)} means that the first formula is used only in the first subnetwork for the location, the second is only used for the scale and the third used for the first and second subnetwork). 

\begin{verbatim}
R> mod <- deepregression(
+    y = y,
+    list_of_formulas = list(
+      location = ~ 1,
+      scale = ~ 1,
+      both = ~ 0 + embd_mod(texts)
+    ),
+    mapping = list(1, 2, 1:2),
+    list_of_deep_models = list(embd_mod = embd_mod), 
+    data = airbnb
+  )
\end{verbatim}  

This example creates an additive model where both the mean and the scale parameter are trained by the same network based on the text information \code{texts}. Following the last layer, the two network outputs extracted from those descriptions are then split and added to the respective additive predictor. 

As an alternative to shared training, we could also define the network with a single output unit and simply add it directly to both parameters. This, however, does not result in the aforementioned reduction in model complexity due to shared weights:

\begin{verbatim}
R> embd_mod <- function(x) x %>%
+    layer_embedding(input_dim = 1000,
+      output_dim = 100) %>%
+    layer_lambda(f = function(x) k_mean(x, axis = 2)) %>%
+    layer_dense(20, activation = "tanh") %>% 
+    layer_dropout(0.3) %>% 
+    layer_dense(1) 

R> form_lists <- list(
+    location = ~ 1 + embd_mod(texts),
+    scale = ~ 1 + embd_mod(texts)
+  )
  
R> mod <- deepregression(
+    y = y,
+    list_of_formulas = form_lists,
+    list_of_deep_models = list(embd_mod = embd_mod), 
+    data = airbnb
+  )
\end{verbatim}

\subsubsection{Optimizer and learning rate}

Per default, \code{deepregression} uses the function \code{keras\_dr} as \code{model\_builder}. The user can supply arguments to \code{keras\_dr} via the ellipsis. In \code{keras\_dr} the \code{optimizer} from \pkg{keras} is defined. It is therefore possible to specify any optimizer in \code{deepregression} that is available in \pkg{keras}, including \code{tf\$keras\$optimizers\$Adadelta},  \code{tf\$keras\$optimizers\$Adam}, \code{tf\$keras\$optimizers\$RMSprop}, or standard stochastic gradient descent without momentum or adaptive learning rates  (\code{tf\$keras\$optimizers\$SGD}). The software uses Adam as the default option. This can be changed, e.g., to Adadelta with learning rate 3 and decay 0.1 by passing the argument \code{optimizer = optimizer\_adadelta(lr = 3, decay = 0.1))} to \code{deepregression}. 

\subsubsection{Orthogonalization} \label{sec:oz}

\begin{figure}
    \centering
            \includegraphics[width=0.95\textwidth]{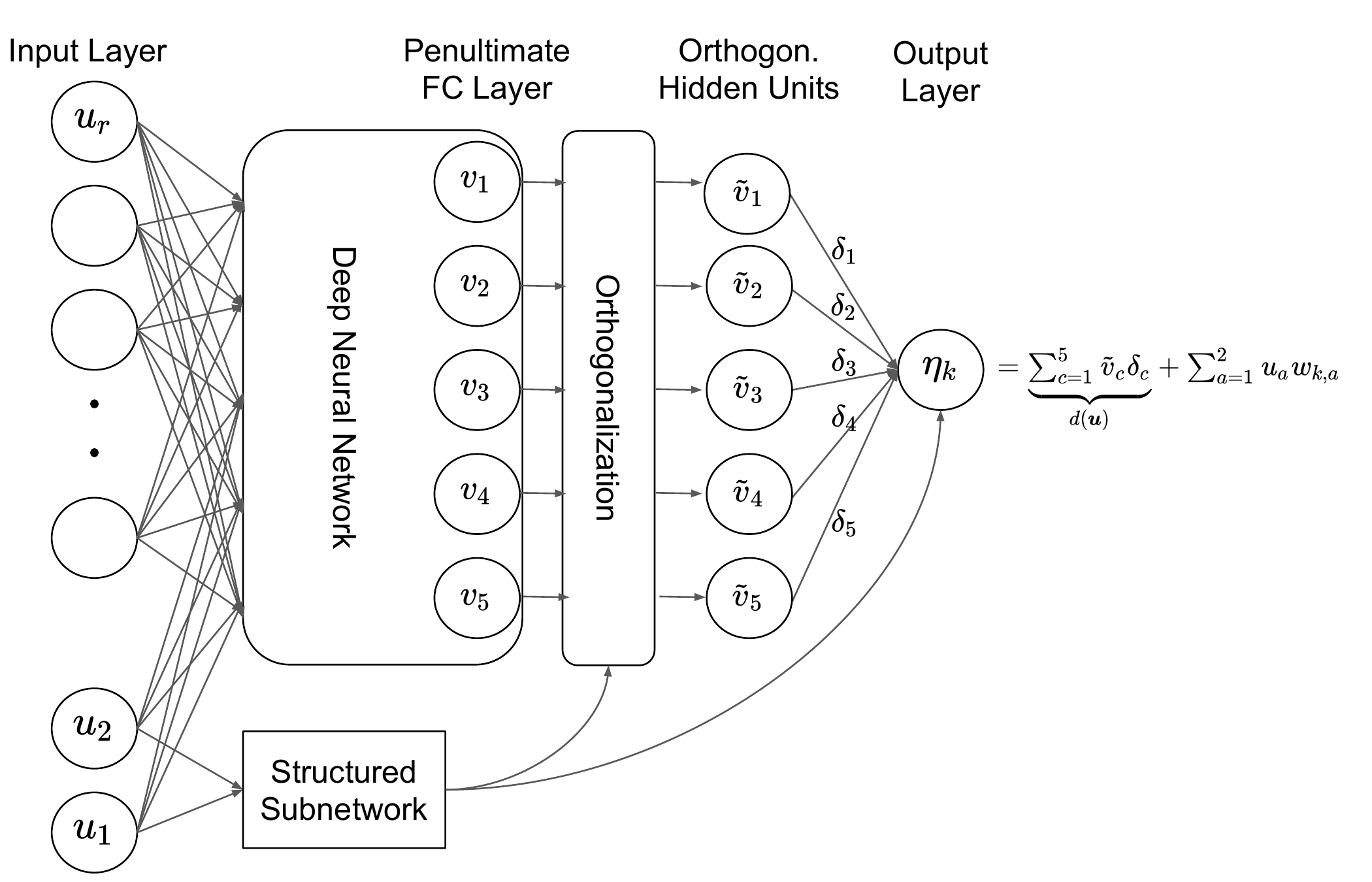}
            \caption{Orthogonalization cell: effects of partially shared  features are learned in a DNN and in a structured subnetwork. The resulting latent features $v_j$ in the DNN are then passed through an orthogonalization cell to ensure identifiability in the predictor $\eta_k$.} \label{fig:oz}
\end{figure}

\code{deepregression} provides options for orthogonalization \citep[for details, we refer the reader to][]{Ruegamer.2020} using the \code{orthog\_options}, which are again defined by a control function \code{orthog\_control}. Per default, \code{orthog\_control} uses the global option to check for orthogonalization, i.e., it defines \code{orthogonalize = options()\$orthogonalize}. This option is set to \code{TRUE} when loading the package and orthogonalizes DNNs in the predictor formulas containing overlapping features with structured terms of the same formula (see Figure~\ref{fig:oz} for a schematic visualization). This ensures identifiability of the structured term(s). Simply put, the orthogonalization constrains the deep network such that its latent learned features cannot represent feature effects that already exist in the structured predictor. Considering, for example, a simple linear model in the structured part, then the orthogonalization would prevent the unstructured (deep) network part from learning the univariate linear feature effects, but not, e.g., their joint interactions of arbitrarily high order. Put slightly more technical, this is achieved by projecting the latent features learned in the unstructured network's penultimate layer onto the orthogonal complement of the column space spanned by the linear features. In a simple linear model, the classic projection or hat matrix $\bm{P}$ projects $\bm{Y}$ onto the smaller subspace $\text{col}(X)$ spanned by the features in $X$, resulting in $\hat{\bm{Y}}=\bm{P}\bm{Y}$. Conversely, the annihilator matrix $(\bm{I}-\bm{P})$ projects $\bm{Y}$ onto the orthogonal complement $\text{col}(X)^{\bot}$ of $\text{col}(X)$, resulting in $\hat{\bm{\varepsilon}}=(\bm{I}-\bm{P})\bm{Y}$. This ensures a unique decomposition $\bm{Y} = \hat{\bm{Y}} + \hat{\bm{\varepsilon}} = \bm{P}\bm{Y} + (\bm{I}-\bm{P})\bm{Y}$ for all $\bm{Y}\in\mathbb{R}^{n}$, because the space of response vectors can be written as an orthogonal decomposition of the two subspaces in terms of a direct sum: $\mathbb{R}^{n}=\text{col}(X) \oplus \text{col}(X)^{\bot}$ (orthogonal decomposition theorem). For a linear model, the unstructured DNN is therefore projected onto the subspace where the $\hat{\bm{\varepsilon}}$ ``live'', so as to extend the space of possible predictions, all without interfering with the linear effects. 

Similar, the argument \code{identify_intercept = options()\$identify_intercept} orthogonalizes the DNN w.r.t. the intercept to make the intercept identifiable. This option is \code{FALSE} per default as the interpretation of an intercept in a semi-structured model is not straightforward.\\ 

\textbf{Case study: Orthogonalization} 

In the following, we introduce a toy example to demonstrate the orthogonalization property of SDDR. We define our data generating process as $Y = 2x + \epsilon, \epsilon \sim \mathcal{N}(0,1)$:

\begin{verbatim}
R> n <- 1000
R> toyX <- rnorm(n)
R> toyY = 2*toyX + rnorm(n)    
\end{verbatim}

We here want to recover the linear effect of \code{toyX} with magnitude $2$ in the presence of a DNN that also takes \code{toyX} as input. We define this DNN as follows:

\begin{verbatim}
R> deep_model <- function(x)
+    {
+      x %>% 
+        layer_dense(units = 100, activation = "relu", use_bias = FALSE) %>%
+        layer_dropout(rate = 0.2) %>%
+        layer_dense(units = 50, activation = "relu") %>%
+        layer_dropout(rate = 0.2) %>%
+        layer_dense(units = 1, activation = "linear")
+    }    
\end{verbatim}

Next, we fit a linear regression with linear effect for \code{toyX}. We pass the variable also to the \code{deep\_model}, one time with orthogonalization and one time without orthogonalization.

\begin{verbatim}
R> forms <- list(loc = ~ -1 + toyX + deep_model(toyX), scale = ~ 1)
R> args <- list(
+    y = toyY, 
+    data = data.frame(toyX = toyX), 
+    list_of_formulas = forms, 
+    list_of_deep_models = list(deep_model = deep_model)
+  )

R> w_oz <- orthog_control(orthogonalize = TRUE)
R> wo_oz <- orthog_control(orthogonalize = FALSE)

R> mod_w_oz <- do.call("deepregression", c(args, list(orthog_options = w_oz)))
R> mod_wo_oz <- do.call("deepregression", c(args, list(orthog_options = wo_oz)))

R> mod_w_oz %>% fit(epochs = 1000, early_stopping = TRUE, verbose = FALSE)
R> mod_wo_oz %>% fit(epochs = 1000, early_stopping = TRUE, verbose = FALSE)

R> cbind(
+    with = c(coef(mod_w_oz, which_param = 1)[[1]]),
+    without = c(coef(mod_wo_oz, which_param = 1)[[1]]),
+    linmod = coef(lm(toyY ~ 0 + toyX))
+  )
         with  without   linmod
toyX 2.002293 0.9065553 2.009948
\end{verbatim}

In the model with orthogonalization, the \code{deep\_model} is orthogonalized w.r.t. \code{toyX} to ensure identifiability of the structured linear term. We thus can recover the linear effect in \code{mod\_w\_oz} despite the presence of the DNN which could have captured a linear (or more complex) effect as well. By default, orthogonalization automatically extracts the terms that overlap in the DNNs and the structured model formula. For expert use, there is also a custom orthogonalization available (see Section~\ref{sec:custom-orthogonalization})

\subsection{Advanced usage} \label{sec:advanced-usage}

\pkg{deepregression} allows for several advanced model specifications and user inputs, briefly described next.






\subsubsection{Offsets}

Several statistical models require an offset to be used in one or more (linear) predictors. These can be specified using an \code{offset(...)} term in the respective formulas.\\

\subsubsection{Constraints}

For smooth effects, further options are available using the \code{penalty_control} function which is passed to the argument \code{penalty_options}. Two important options are inherited by \pkg{mgcv}. \code{absorb\_cons} (default \code{FALSE}) will absorb identifiability constraints into the smoothing basis (see \code{?mgcv::smoothCon}) and \code{zero_constraint_for_smooths} will constrain all smooth to sum to zero over their domain, which is usually recommended to prevent identifiability issues (default \code{TRUE}).

\subsubsection{Custom distribution function}

It is also possible to define custom distribution functions to be learned in \code{deepregression} using the \code{create\_family} function and passing the result to \code{deepregression}'s \code{family} argument. To create a custom distribution, a distribution \code{tfd\_dist} from \pkg{tfprobability} must be passed to \code{create\_family}, as well as a list of transformations \code{trafo\_list} defining the different $h_k$s of the distribution. For example, a normal distribution with different response functions $h_1(\eta) = 1/\eta$, $h_2(\eta) = \eta^2$ can be defined by passing the following to the \code{family} argument.

\begin{verbatim}
create_family(
  tfd_dist = tfd_normal,
  trafo_list = list(
    function(x) tf$math$reciprocal(x),
    function(x) tf$math$square(x)
  )
)
\end{verbatim}

\subsection{Custom orthogonalization} \label{sec:custom-orthogonalization}

If there is reason to orthogonalize a DNN w.r.t. a structured model term that is not explicitly present, the \code{\%OZ\%}-operator can be used. On the left side of the \code{\%OZ\%}-operator in the formula, the DNN must be given. On the right side of the operator, either a single structured term (such as \code{x}, \code{s(x)}, \code{te(x,z)}) or a combination of structured terms using brackets around all terms and separated with a \code{+} between each term must be supplied (e.g., \code{deep\_model(u) \%OZ\% (x + s(z))} to project the \code{deep\_model} into the orthogonal complement of the space spanned by the linear effect of \code{x} and the basis of \code{s(z)}).\\ 

\textbf{Case study: Custom orthogonalization} 

We use the previous toy example and pretend that there is another feature that can potentially influence the linear effect of \code{toyX} with the same information content but another name (so that it is not recognized automatically as a candidate for orthogonalization).

\begin{verbatim}
R> toyXinDisguise <- toyX
\end{verbatim}

\begin{verbatim}
R> form_known <- list(loc = ~ -1 + toyX + deep_model(toyX), scale = ~ 1)
R> form_unknown <- list(
+    loc = ~ -1 + toyX + deep_model(toyXinDisguise), 
+    scale = ~ 1
+  )
R> form_manual <- list(
+    loc = ~ -1 + toyX + 
+      deep_model(toyXinDisguise) %OZ% (toyXinDisguise), 
+    scale = ~ 1
+  )
R> args <- list(
+    y = toyY, 
+    data = data.frame(toyX = toyX, toyXinDisguise = toyXinDisguise), 
+    list_of_deep_models = list(deep_model = deep_model)
+  )

R> mod_known <- do.call(
+    "deepregression", c(args, list(list_of_formulas = form_known))
+  )
Preparing additive formula(e)... Done.

R> mod_unknown <- do.call(
+    "deepregression", c(args, list(list_of_formulas = form_unknown))
+  )
Preparing additive formula(e)... Done.

R> mod_manual <- do.call(
+    "deepregression", c(args, list(list_of_formulas = form_manual))
+  )
Preparing additive formula(e)... Done.

R> mod_known %>% fit(epochs = 1000, early_stopping = TRUE, verbose = FALSE)
R> mod_unknown %>% fit(epochs = 1000, early_stopping = TRUE, verbose = FALSE)
R> mod_manual %>% fit(epochs = 1000, early_stopping = TRUE, verbose = FALSE)

R> cbind(
+    known = coef(mod_known, params = 1)[[1]],
+    unknown = coef(mod_unknown, params = 1)[[1]],
+    manual = coef(mod_manual, params = 1)[[1]]
+  )
         [,1]     [,2]     [,3]
[1,] 2.053214 1.744017 2.054634
\end{verbatim}

\subsection{Working with images}

The DNNs in SDDR can also incorporate images into the additive predictor of one or more parameters. Due to their size, images are usually not loaded into memory completely, but read in mini-batches from disk during training. For this purpose, \pkg{deepregression} offers a generator that creates small batches of images and other tabular information for training and prediction. The absolute path to the images must be given as string in the \code{data}. In addition, the user must then specify the image size in the respective element in the \code{list\_of\_deep\_models} by providing a list with the name of the DNN and its size.\\

Some care needs to be taken with respect to interpreting the effect of unstructured data modalities, such as images, on the predictors $\eta_k, k=1,\ldots,K$, and whether they affect the interpretation of structured linear coefficients. At a basic mathematical level, images are typically represented as rank 3 tensors of shape $(\text{height}, \text{width}, \text{channels})$, i.e., a collection of non-negative numbers endowed with a spatial structure not unlike other constructs studied in statistics, e.g., functional or image data in a general functional regression context that also permits scalar covariates \citep[see, e.g.,][]{greven2017general}. The additive structure of of the predictor $\eta_k$ in equation (1) in principle enables a classic \textit{ceteris paribus} comparison, thereby leaving the interpretation of the linear coefficients unchanged, bar the fact that they are now also conditioned on the image. A unit change in the linear feature $x_j$ yields a $\beta_j$ change in $\eta_k$ for a given image, and by exchanging the images, we can calculate differences in $\eta_k$ to evaluate the effect of moving around the pixel space of images. Although one can often not interpret individual pixels \citep[only if the images are preregistered, aligned and show the same underlying scene as, e.g., done in][]{lemhadri2021lassonet}, we can make use of the representation learning property of modern CNNs. This allows partially interpretable comparisons at very high level learned features, even though the pixels from which the latent features were learned have no natural interpretation.\\

\textbf{Case study: Working with images}

The pictures of each of the apartments in the \code{airbnb} data can be downloaded from \url{https://github.com/davidruegamer/airbnb}. Here, the folder \code{data/pictures/32} contains all pictures from Munich. Assuming the pictures are stored in the home directory in a dedicated folder called \code{airbnb}, we first create the absolute path as follows:

\begin{verbatim}
R> airbnb$image <- paste0("/home/user/airbnb/data/pictures/32/",
+    airbnb$id, ".jpg")
\end{verbatim}

Next, we define an appropriate DNN architecture to extract latent features from the images. Here, we use a convolutional neural network (CNN). First, we define a CNN block:

\begin{verbatim}
R> cnn_block <- function(
+    filters, 
+    kernel_size, 
+    pool_size, 
+    rate, 
+    input_shape = NULL
+    ){
+      function(x){
+        x %>% 
+        layer_conv_2d(filters, kernel_size, 
+          padding="same", input_shape = input_shape) %>% 
+        layer_activation(activation = "relu") %>% 
+        layer_batch_normalization() %>% 
+        layer_max_pooling_2d(pool_size = pool_size) %>% 
+        layer_dropout(rate = rate)
+      }
+    }
\end{verbatim}

We  then create the DNN function required by \pkg{deepregression} using a single block:

\begin{verbatim}
R> cnn <- cnn_block(
+    filters = 16, 
+    kernel_size = c(3,3), 
+    pool_size = c(3,3), 
+    rate = 0.25,
+    shape(200, 200, 3)
+  )
R> deep_model_cnn <- function(x){
+    x %>% 
+    cnn() %>%
+      layer_flatten() %>% 
+      layer_dense(32) %>% 
+      layer_activation(activation = "relu") %>% 
+      layer_batch_normalization() %>% 
+      layer_dropout(rate = 0.5) %>% 
+      layer_dense(1)
+  }
\end{verbatim}

Note that we have to provide the shape of the image in the first layer of the CNN, which is $200 \times 200$ with three colour channels. Finally, we define an illustrative \code{deepregression} model using an additive predictor for the mean with both, structured effects for, e.g., the room type as linear effect, and an unstructured effect by including the image information:

\begin{verbatim}
R> mod_cnn <- deepregression(
+    y = y,
+    list_of_formulas = list(
+      ~1 + room_type + bedrooms + beds + 
+        deep_model_cnn(image), 
+      ~1 + room_type), 
+    data = airbnb,
+    list_of_deep_models = list(deep_model_cnn = 
+      list(deep_model_cnn, c(200,200,3))),
+    optimizer = optimizer_adam(lr = 0.0001)
+  )
\end{verbatim}

The training using images is usually slower, but also requires less iterations. For the given example, we can, e.g., use the following setting for model fitting:

\begin{verbatim}
R> mod_cnn %>% fit(
+    epochs = 100, 
+    early_stopping = TRUE, 
+    patience = 5,
+    verbose = FALSE
+  )
\end{verbatim}

Checking the influence of the image in additive predictors is not straightforward. One strategy is to set all other tabular variables to zero or to the reference category and to qualitatively inspect predictions of the SDDR network. 

\subsection{Uncertainty quantification}

Next to analyzing prediction performance and the estimated partial effects, a crucial step in evaluating the model's reasoning is to assess the uncertainty of the model. While distributional regression directly models the uncertainty in the data generating process (the so-called \emph{aleatoric uncertainty}), the uncertainty of the model (\emph{epistemic uncertainty}) is of vital importance as well. \pkg{deepregression} provides two ways to characterize the variability of model estimates: 1) in the function space, i.e., the predicted outcome space, and 2) in the weight space for structured components, i.e., statistical inference for the weights of the structured model part.

\subsubsection{Function space uncertainty}

The solutions obtained by optimizing neural networks depend heavily on the stochastic 
nature of the optimization algorithm. For instance, re-running the same optimization 
on the same data yields different models with heterogeneous test performance due 
to the stochasticity of the optimization routine and different weight initializations.
This function space uncertainty is taken advantage of in \emph{deep ensembles} 
\citep{lakshminarayanan2016deepensembles} by constructing an ensemble of neural networks.
Deep ensembles linearly pool the predictions of its members, which, in many cases, improves prediction performance and leads to better out-of-distribution generalization and 
uncertainty quantification. 
Pooling the ensemble members can be seen as an approximate Bayesian method,  
explaining their purported benefits in epistemic uncertainty estimation 
\citep{maddox2019simple,wilson2020bayesian}.

In \pkg{deepregression}, deep ensembles are implemented as a mixture distribution 
of the predicted distributions of its members and can be trained using the 
\code{ensemble()} function. The resulting distribution is a \code{tfd_distribution}
object, which allows numerous downstream computations, including the log-likelihood,
cumulative distribution and survivor function.\\

\textbf{Case study: Quantifying predictive uncertainty} 

The following fits a deep ensemble for the Airbnb data using early stopping and a
20\% validation split.

\begin{verbatim}
R> mod <- deepregression(
+    y = y, 
+    data = airbnb,
+    list_of_formulas = list(
+      location = ~ 1 + s(longitude),
+      scale = ~ 1 + s(longitude)
+    )
+  )
R> ens <- ensemble(mod, n_ensemble = 5, epochs = 1e3, early_stopping = TRUE,
+     validation_split = 0.2)
R> ensemble_distr <- get_ensemble_distribution(ens, data = newdata)
\end{verbatim}

\begin{figure}[!ht]
    \centering
    \includegraphics[width=0.49\textwidth]{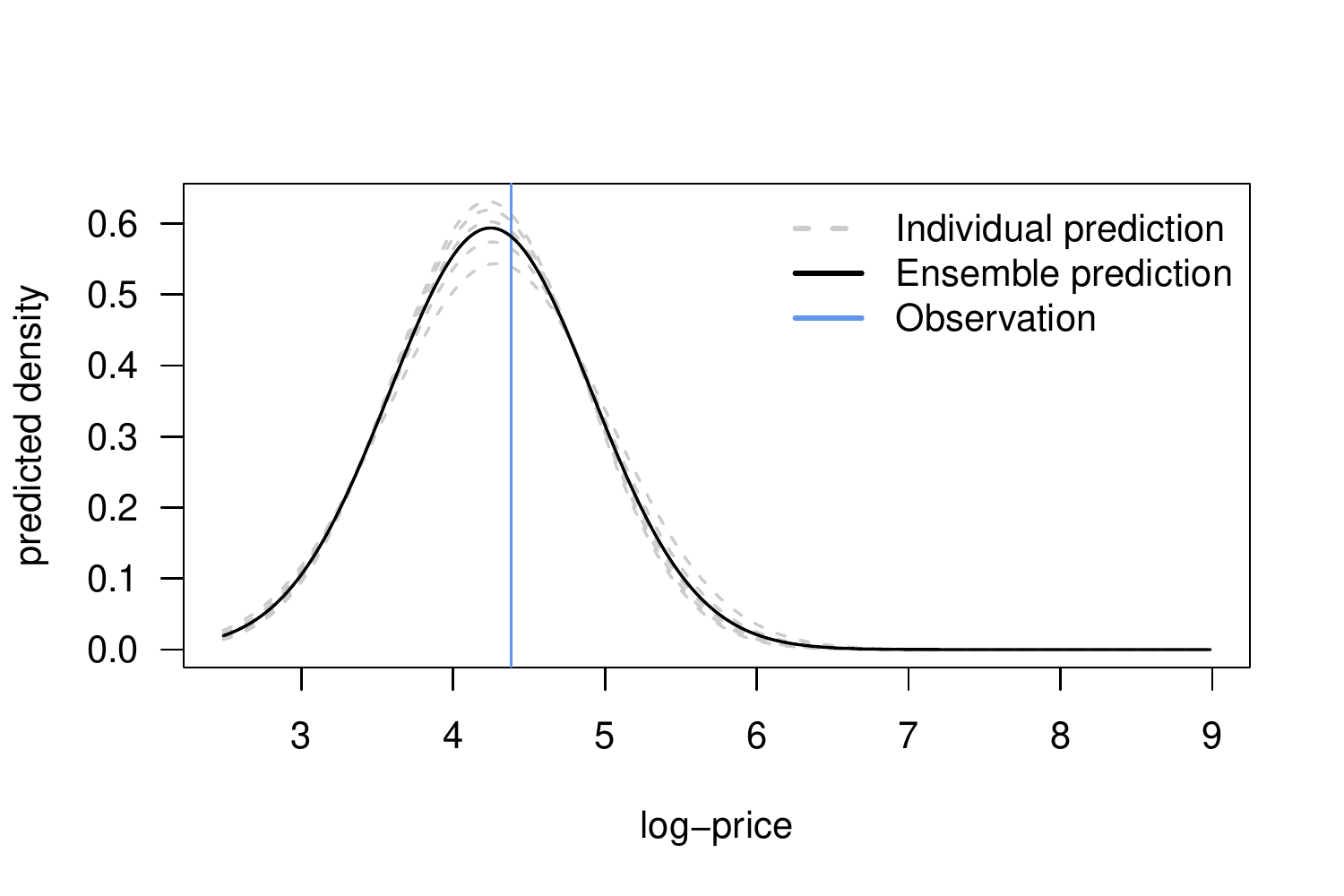}
    \includegraphics[width=0.49\textwidth]{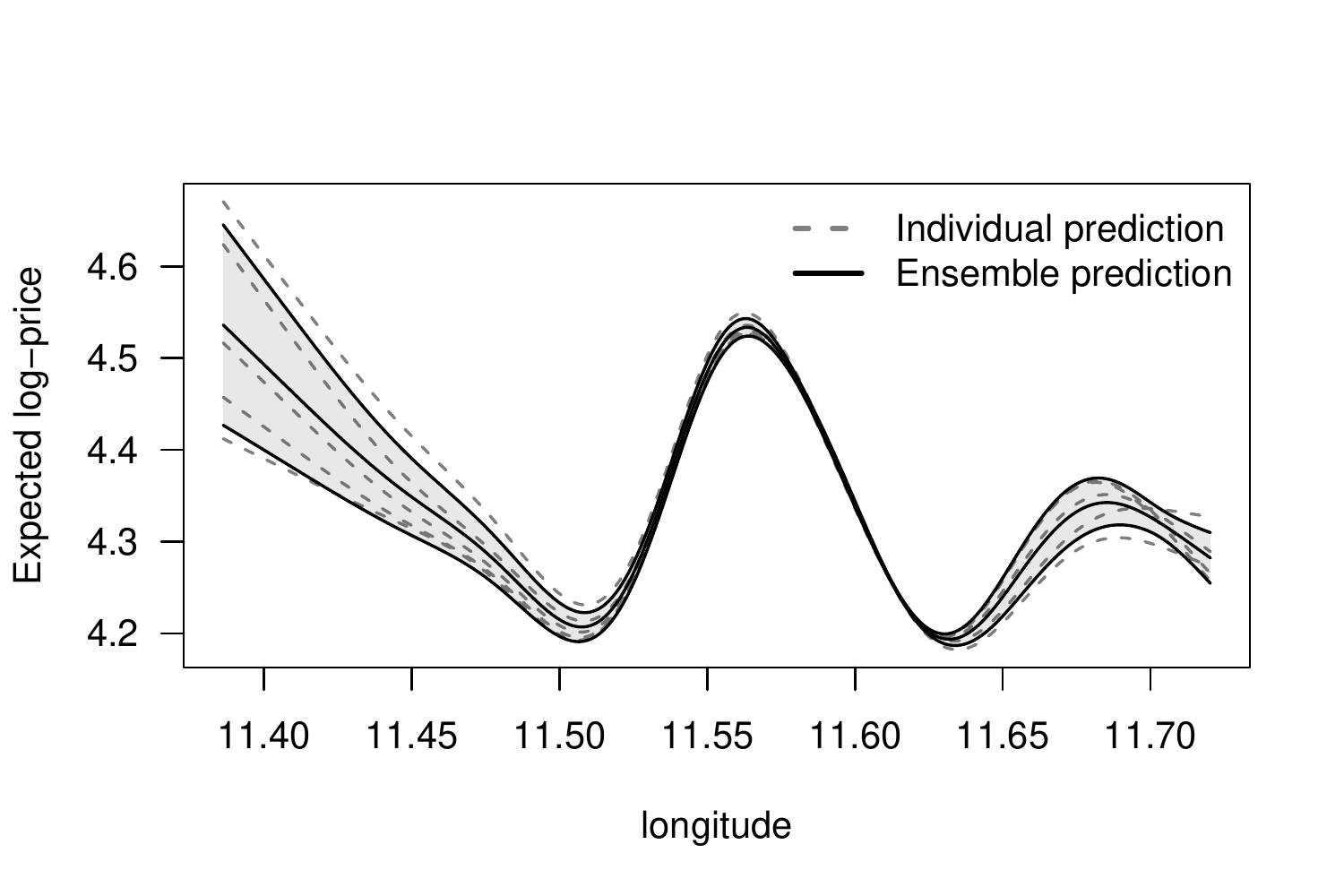}
    \caption{Left: Individual and combined \code{deepregression} models for the 
    predicted conditional distribution of a single observation with observed log-price 
    visualized by a vertical line. Individual predicted densities of the ensemble members 
    are shown as dashed lines, and the ensemble prediction as a solid line.
    The average and ensemble negative log-likelihood (NLL) for the single observation 
    are shown on the right. Right: Expected log-price as a function of longitude with 
    epistemic uncertainty ($\pm$ point-wise standard deviation of the ensemble members).}
    \label{fig:ensemble}
\end{figure}

Figure~\ref{fig:ensemble} (left) depicts the predicted rating density of a single 
observation for all ensemble members (dashed lines) as well as the ensemble itself 
(solid line), together with the actually observed rating and the negative log-likelihood 
for that observation. The ensemble reduces the influence of overconfident predictions 
and improves upon the average loss of the single models.
The right panel illustrates the epistemic uncertainty quantification by showing the 
estimated expected logarithmic apartment price as a function of longitude.

\subsubsection{Weight space uncertainty}

Next to uncertainty quantification in the function space, the uncertainty in the weight space of the neural network is not of lesser importance. While it is difficult to interpret or infer insights from uncertainties of weights in the deep network part(s), the uncertainty in the structured part is what more classical regression approaches refer to as statistical inference. Since deep neural network components in the additive predictors can usually be represented as $d_{k,l}(\bm{u}) = \tilde{\bm{u}}_{k,l}^\top \tilde{\bm{w}}_{k,l}$ with latent features $\tilde{\bm{u}}_{k,l}$ and weights $\tilde{\bm{w}}_{k,l}$, the final additive predictor in semi-structured regression can also be seen as an extended structured predictor with random features $\tilde{\bm{u}}_{k,l}$. One possibility to quantify the uncertainty of the structured part is hence to extract the latent features of deep neural network components and refit the last layer of the model. This is similar to a local linearization approach \citep{immer2021improving} and referred to as \emph{last layer inference} \citep{daxberger2021laplace}. Note, however, that this approach underestimates the variance in model as it neglects the uncertainty of previous layers in the deep network components. Similar to deep ensembles, repeating network training and refitting might give thus additional insights into how much this variance influences structured network components.\\

\textbf{Case study: Last layer statistical inference} 

The following example fits a simple example of a deep generalized additive model used in the beginning of this article and then refits the model based on the latent features learned in the last layer of the deep neural network. We first define the neural network as follows:

\begin{verbatim}
R> deep_model <- function(x)
+    {
+       x %>% 
+       layer_dense(units = 5, activation = "relu", use_bias = FALSE) %>%
+       layer_dense(units = 3, name = "penultimate_layer") %>%
+       layer_dense(units = 1, activation = "linear")
+    }
\end{verbatim}

Note that we can name the penultimate layer to make it easier later to access the latent features of this layer. We then define and fit the model.

\begin{verbatim}
R> mod <- deepregression(
+    y = y, 
+    data = airbnb,
+    list_of_formulas = list(
+      location = ~ 1 + beds + s(accommodates, bs = "ps") +
+         s(days_since_last_review, bs = "tp") + 
+         deep(review_scores_rating, reviews_per_month),
+      scale = ~1
+    ),
+    list_of_deep_models = list(deep = deep_model)
+  )
R> mod %>% fit(
+    epochs = 100, 
+    verbose = FALSE, 
+    view_metrics = FALSE,
+    validation_split = 0.2
+  )
\end{verbatim}

To extract the learned latent features of the model, we create a temporary keras model to extract the intermediate predictions $\tilde{\bm{u}}$ (here 3-dimensional):

\begin{verbatim}
R> intermediate_mod <- keras_model(mod$model$input, mod$model$get_layer(
+    name="penultimate_layer")$output
+  )
R> newdata_processed <- prepare_newdata(
+    mod$init_params$parsed_formulas_contents, 
+    airbnb, 
+    gamdata = mod$init_params$gamdata$data_trafos
+  )
R> tilde_u <- as.data.frame(intermediate_mod$predict(newdata_processed))
str(tilde_u, 1)
'data.frame':	3504 obs. of  3 variables:
 $ V1: num  52.7 53.7 49.5 53.2 52.7 ...
 $ V2: num  -21.4 -21.7 -20.1 -21.5 -21.3 ...
 $ V3: num  34.7 34.3 32.9 34.4 34.2 ...
\end{verbatim}

\begin{figure}[!t]
    \centering
    \includegraphics[width=0.9\textwidth]{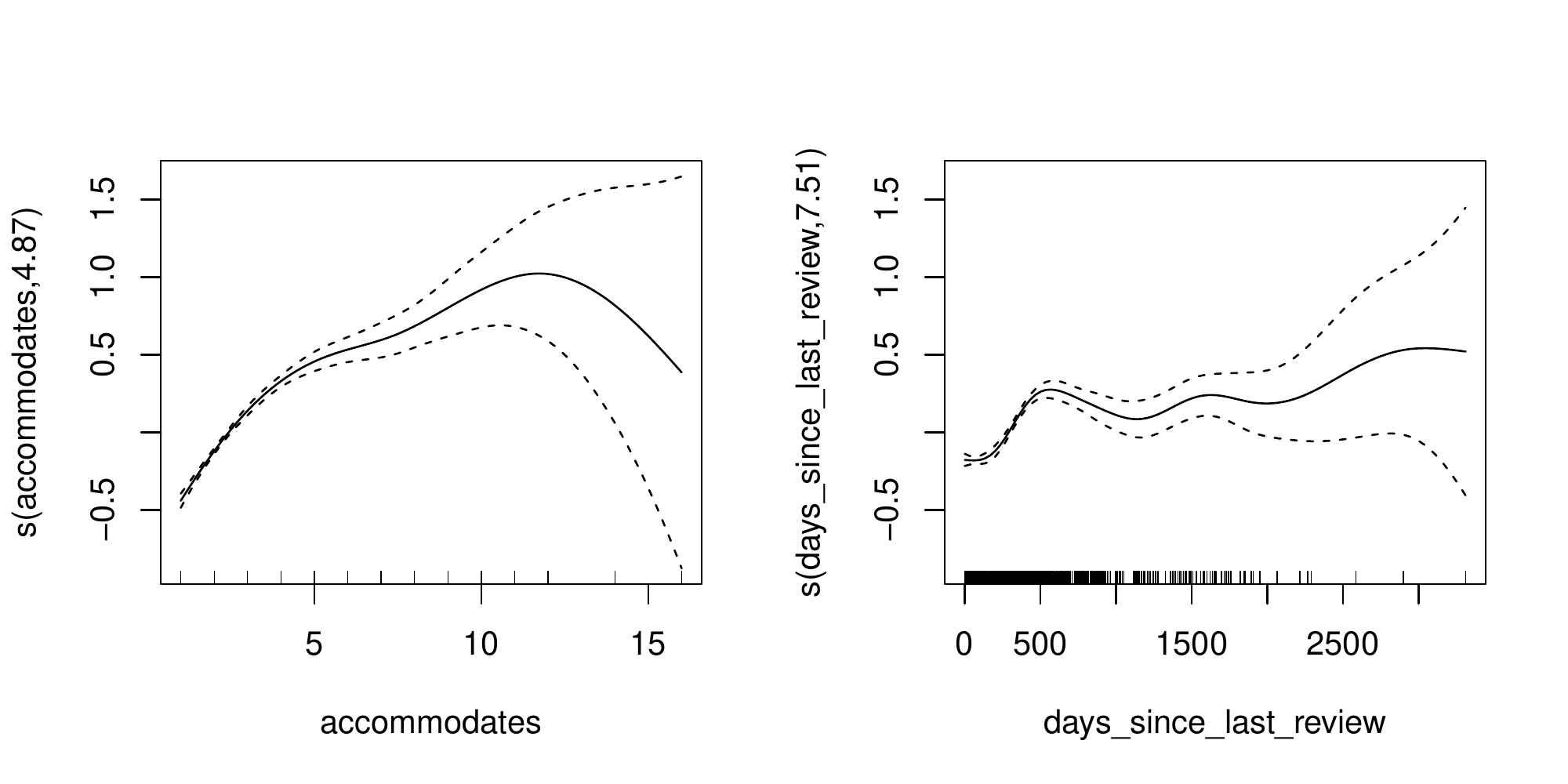}
    \caption{Estimated non-linear effects and their uncertainty obtained from the refitted Deep GAM.}
    \label{fig:uqweights}
\end{figure}

Given \code{tilde\_u}, we now refit the last layer using \pkg{mgcv} and extract the uncertainty statements of interest (here confidence intervals for non-linear effects depicted in Figure~\ref{fig:uqweights}).

\begin{verbatim}
R> gam_mod <- gam(log(price) ~ 
+    1 + beds + s(accommodates, bs = "ps") + 
+    s(days_since_last_review, bs = "tp") + 
+    V1 + V2 + V3,
+    data = cbind(as.data.frame(airbnb), tilde_u)
+  )
R> plot(gam_mod, pages = 1)
\end{verbatim}

\section{Conclusion}\label{sec:conclusion} 
We have presented \pkg{deepregression}, a software package that implements the SDDR framework in \proglang{R}. \pkg{deepregression} allows to learn the parameters of a wide range of distributions based on a combination of additive regression models and deep neural networks. The implementation is comprised of three essential components:
(1) a modular neural network building system for the seamless combination of statistical and deep learning approaches, (2) an orthogonalization cell (as put forward by \citealp{Ruegamer.2020}) to allow for an interpretable combination of different subnetworks, and (3) comprehensive pre-processing tools to set up the models. The software provides a common formula interface to specify distributional parameters, allows to learn shared weights and automatically handles identifiability issues. We believe that \pkg{deepregression}'s modular design and functionality fosters rapid and reproducible prototyping of complex (combinations of) statistical and deep learning models in one common environment.      


\section*{Acknowledgements}
This work has been partially supported by the German Federal Ministry of Education and Research (BMBF) under Grant No. 01IS18036A. The authors of this work take full responsibilities for its content.
Nadja Klein gratefully acknowledges  support  by the German research foundation (DFG) through the  Emmy Noether grant KL 3037/1-1. We would also like to thank Matthias Schmid and the anonymous reviewers for their suggestions and comments, which have been very helpful in improving the manuscript.

\bibliography{bibliography}

\end{document}